\documentclass[11pt]{article}
\usepackage[natbib, preprint]{jmlr2e}
\usepackage{subcaption}
\usepackage[utf8]{inputenc} 
\usepackage[T1]{fontenc}    
\usepackage{hyperref}       
\usepackage{url}            
\usepackage{booktabs}       
\usepackage{amsfonts}       
\usepackage{nicefrac}       
\usepackage{microtype}      
\usepackage{tabularx}
\usepackage{graphicx}
\usepackage{amsfonts}
\usepackage{tikz}
\usepackage{amsmath}
\usepackage{cleveref}
\usepackage[alpha]{mdpn}
\usepackage{comment}
\usepackage[rightcaption]{sidecap}
\usetikzlibrary{arrows.meta}
\usepackage{algorithm}
\usepackage[noend]{algpseudocode}
\makeatletter
\def\BState{\State\hskip-\ALG@thistlm}
\makeatother
\newcommand\numberthis{\addtocounter{equation}{1}\tag{\theequation}}
\newcommand*\samethanks[1][\value{footnote}]{\footnotemark[#1]}
\newcommand\independent{\protect\mathpalette{\protect\independenT}{\perp}}
\def\independenT#1#2{\mathrel{\rlap{$#1#2$}\mkern2mu{#1#2}}}

\begin{document}
\title{Causality and Batch Reinforcement Learning:\\ Complementary Approaches To Planning In Unknown Domains}
\author{\name  James Bannon\thanks{Equal contribution }\email{jjb509@nyu.edu}  \\
       \name Wenbo Song\samethanks\email{ws1542@nyu.edu}\\
       \name Brad Windsor\samethanks \email{bw1879@nyu.edu}\\
       \addr{Courant Institute of Mathematical Sciences,\\
       New York University, New York, NY, 10012}
        \AND
       \name Tao Li\samethanks \email{tl2636@nyu.edu} \\
       \addr{Department of Electrical and Computer Engineering,\\
       New York University, New York, NY, 11201}
       }

\maketitle
\begin{abstract}
   Reinforcement learning algorithms have had tremendous successes in online learning settings. However, these successes have relied on low-stakes interactions between the algorithmic agent and its environment. In many settings where RL could be of use, such as health care and autonomous driving, the mistakes made by most online RL algorithms during early training come with unacceptable costs. These settings require developing reinforcement learning algorithms that can operate in the so-called {\em batch} setting, where the algorithms must learn from set of data that is fixed, finite, and generated from some (possibly unknown) policy. Evaluating policies different from the one that collected the data is called {\em off-policy evaluation}, and naturally poses {\em counter-factual} questions. In this project we show how off-policy evaluation and the estimation of treatment effects in causal inference are two approaches to the same problem, and compare recent progress in these two areas. 
\end{abstract}

\section{Introduction}\label{sec:intro}
\paragraph{The Limits of Limitless Data} Reinforcement learning (RL) distinguishes itself from supervised and unsupervised learning by being interactive \citep{sutton2018reinforcement}. In the typical RL setting, an agent interacts with its environment in an ongoing way with the goal of learning a {\em policy} that is optimal in some sense. Theoretical analyses of these algorithms tend to fall into two categories. The first estimates long-term optimality by bounding the expected {\em regret} of the algorithm, that is the difference between the algorithms' actual choices and those that would have been optimal. The second looks to demonstrate on either the long-run convergence to a truly optimal policy or estimate the expected {\em return} of a policy and subsequently show it is optimal. 

Powered by advances in deep learning and computing power, we have seen many great successes in reinforcement learning, a very general learning paradigm that can
model a wide range of problems, such as games \citep{mnih2015human, silver2016mastering, tesauro1995temporal,jaderberg2019human, vinyals2019grandmaster}, robotics \citep{kober2013reinforcement}, autonomous driving \citep{sallab2017deep}, healthcare \citep{komorowski2018artificial}, and many others. As pointed out in \citep{li2019perspective}, the quality of a policy is often measured by the average reward received if the policy is followed by the agent to select actions. That is, many successful reinforcement learning applications rely on on-policy data collected through online learning. High profile success of RL, most of which are in the field of super-human level game playing, have been done in this setting where the only cost of algorithmic error is additional computation. Put differently, early mistakes made by the algorithm before (asymptotic) convergence only require that the agent be restarted and have no consequences in the physical world. Many of the aforementioned successes leverage this low-stakes setting by running for several million iterations before convergence. 

This dependence on the ability to acquire an arbitrarily large amount of data proves to be a strong limitation when attempting to apply these algorithms to certain real-world scenarios. In domains like healthcare, autonomous driving, or automated plant control, the costs of making mistakes are unacceptably high, and collecting millions of data points is infeasible. Instead, reinforcement learning algorithms must proceed from a fixed set of data 
which may be gathered by a collection of (possibly unknown) policies. The challenges to batch RL algorithms come primarily from the fact that they must do {\em off-policy evaluation}, that is they need to measure the performance  of some policy that may differ arbitrarily from the policies used to collect the data. In doing this an implicitly counterfactual question is asked: ``what would the value of the total reward have been if the algorithm had performed $a'$ instead of $a$?''

In this paper, we argue that causal inference and off-policy RL are two different approaches to similar (and sometimes  identical) questions, and that the fields can benefit from mutual dialogue. 
In \cref{sec:caus} we describe the field of causality, and formalize two settings in which causal inference is often done: the potential outcomes framework and structural causal models. Then in \cref{sec:bandits} we give connections between these causal frameworks and sequential decision making by focusing on causality in contextual bandit problems, which paves the way for our discussion about causality in batch RL and off-policy evaluation.  In \cref{sec:batchRL}, we provide an overview of batch RL and off-policy evaluation and point out failures of these algorithms to generalize well to unseen data. In \cref{sec:OPPE}, we demonstrate that causal approaches address weaknesses of these algorithms, leading to estimators with better generalization in off-policy evaluation. Section 6 links two final subdomains of these fields. Lastly in \cref{sec:conclusion} we summarise our work and suggest avenues for study. 

\section{Causality}\label{sec:caus}

Causality is best understood as the science of predicting outcomes from {\em interventions}. This sets it apart from typical supervised machine learning (ML) problems where the chief concern is making predictions based on {\em observations.} Most ML algorithms are designed to use a labeled data set $\{(X_i,Y_i)\}_{i=1}^n$ to learn a function $f:X\to Y$ which can then be used to make predictions $\hat{Y}=f(X_{new})$ on new data items $X_{new}$. Causality, on the other hand, cares about designing algorithms for predicting $Y$ when $X$ is {\em actively manipulated}. 

The distinction arises from the fact that observing $X$ does not include the latent influences that lead to that observation. Consider the following example: suppose data $\{(X_i,Y_i)\}$ is collected on a population where $X_i\in\{0,1\}$ represents a person taking a specific multi-vitamin and $Y_i\in\mathbb{R}_{\geq0}$ is their average daily cortisol levels (a measure of stress). For the $i$'th person, predicting $Y_i$ from observing $X_i$ ignores the processes that lead to observing that $X_i$ occurred. If, for example, people with a certain mindfulness meditation app are more likely to take the multi-vitamin (perhaps the app advertises it) and the app itself is what lowers stress levels. In this case any positive association between the multi-vitamin and lower stress will be {\em confounded} with the presence of the app being an {\em unobserved confounder.} This scenario is depicted in \Cref{fig:causalmodel} and described in more detail in \cref{ssec:scm} below.

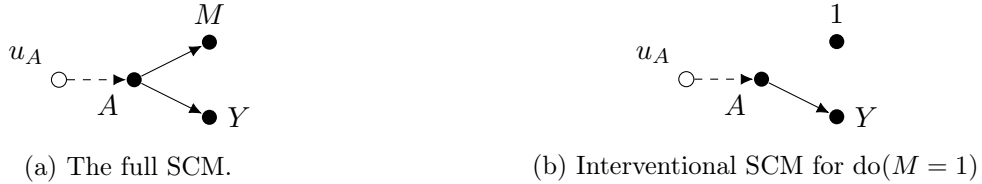
\begin{figure}[ht]
\begin{subfigure}{.5\textwidth}
  \centering
\begin{tikzpicture}
    \node[circle,fill=none,draw,minimum size=2mm](u_A) at (-2,0)[circle,fill=none,draw,inner sep=1pt,label= above left:$u_A$]{};
    \node[fill,circle,minimum size=2mm] (M) at (0,0.5)[circle,fill,inner sep=1pt,label= above:{$M$}]{};
    \node[fill,circle,minimum size=2mm] (A) at (-1,0)[circle,fill,inner sep=1pt,label= below left:$A$]{};
    \node[fill,circle,minimum size=2mm] (Y) at (0,-0.5)[circle,fill,inner sep=1pt,label=right:$Y$]{};
    \draw[-{Latex[width=1.5mm]}](A)--(Y);
   \draw[-{Latex[width=1.5mm]}] (A)--(M);
    \draw[-{Latex[width=1.5mm]},dashed] (u_A)--(A);
\end{tikzpicture}
  \caption{The full SCM.}
  \label{fig:causalmodel}
\end{subfigure}
\begin{subfigure}{.5\textwidth}
  \centering
\begin{tikzpicture}
     \node[circle,fill=none,draw,minimum size=2mm](u_A) at (-2,0)[circle,fill=none,draw,inner sep=1pt,label= above left:$u_A$]{};
    \node[fill,circle,minimum size=2mm] (M) at (0,0.5)[circle,fill,inner sep=1pt,label= above:{$1$}]{};
    \node[fill,circle,minimum size=2mm] (A) at (-1,0)[circle,fill,inner sep=1pt,label= below left:$A$]{};
    \node[fill,circle,minimum size=2mm] (Y) at (0,-0.5)[circle,fill,inner sep=1pt,label=right:$Y$]{};
    \draw[-{Latex[width=1.5mm]}](A)--(Y);
    \draw[-{Latex[width=1.5mm]},dashed] (u_A)--(A);
\end{tikzpicture}
  \caption{Interventional SCM for $\text{do}(M=1)$}
  \label{fig:intervention}
\end{subfigure}
\caption{Depiction of intervention in a SCM. Empty nodes are random variables in the set $U$ where solid nodes are observed/measured variables in the set $V$. Dashed lines are from unobserved variables to observed ones. Solid lines are between observed variables. $A$ represents the app, $M$ the multi-vitamin, and $Y$ the cortisol levels.}
\label{fig:causal_example}
\end{figure} 
Thus we may be able to {\em predict} the cortisol levels given that someone takes the multi-vitamin, but we won't be picking up on a true mechanistic cause. If the data set is collected from a population that mostly has the mobile app then it will appear that there is a positive association between taking the vitamin and lowered cortisol level. If new samples are drawn from that same population it is possible to predict cortisol levels with a high degree of accuracy (low loss) given just the fact that they take a multi-vitamin, but this won't provide any information about {\em how} this influence occurs. Put differently, just because we can {\em predict} the cortisol levels from the fact that someone takes the multi-vitamin, doesn't mean we can tell what will happen to someone's cortisol if we {\em force them to take the multi-vitamin.}

Underpinning this issue is that $X_i$ and $Y_i$ are not independent. Ideally they would be unconfounded, which happens if we can measure a set of variables such that $X_i$ and $Y_i$ are conditionally independent given those variables. This reasoning underpins the randomization behind clinical trials in pharmaceutical development. If the {\em treatment} (in our example the multi-vitamin) is randomized, then any confound between the treatment and the outcome is eliminated. Obviously, clinical trials and randomization are a luxury and most causal questions must be asked of {\em observational } data, which is where most of the challenge in causal inference arises.

In the causality literature there are two distinct but symbiotic kinds of causal questions. The first kind of question is {\bf causal discovery}. Causal discovery algorithms are given a collection of observations of variables $X_1, \dots, X_k$ and try to learn causal relationships between them. The difficulty of this problem depends strongly on what is meant by a causal relationship. For example, if causality is just conditional dependence then the task is equivalent to learning Bayesian network structure, which is known to be  NP-hard \citep{chickering2004large}. Other views, such as those in \citep{kleinberg2012temporal}, can be learned in polynomial time. 

The second kind of question is {\bf causal inference.} In causal inference, hypothesized causal relationships between covariates are assumed and the subsequent task is to infer {\em the degree to which one has a causal effect on another}. This is another distinction between the causal inference and supervised machine learning. In the latter setting we care about predictive performance, in the former the properties of our estimators are what is important. For the rest of this project, we abandon the question of causal discovery and focus on causal inference as it is applied to batch reinforcement learning. 

In the two subsections below we review two causality frameworks for carrying out causal inference. The first, the potential outcomes framework, has a long history in statistics. The second, structural causal models, is a more general framework that is a recent outgrowth of the computer science, economics, and graphical models fields. In both cases we discuss how causal effects can be estimated. 

\subsection{The Potential Outcomes Framework}

The potential outcomes framework (sometimes called the Rubins Causal Model (RCM)) \citep{sekhon2008neyman,rubin2005causal,splawa1990application} has its roots in the statistics literature. The setting involves an observational data set of the form $\{(Y_i, t_i, x_i)\}$ where for the $i$'th person $Y_i$ is an outcome of interest (such as cortisol levels), $t_i\in\{0,1\}$ is a variable indicating treatment (e.g. taking the multi-vitamin), and $x_i$ is a vector of covariates. The RCM postulates the existence of a joint distribution of {\em potential outcomes} $(Y_1, Y_0)$ corresponding, respectively, to the case where a person is or is not treated. We use the notation $Y_j(x_i), j\in\{0,1\}$ to denote the potential outcome for a specific set of covariates with and without treatment.

A common assumption in this framework is that of {\em strong ignorability}, which can be written symbolically as
\begin{align}
    (Y_1, Y_0)\independent{}{}t|x & & 0<Pr(t=1|x)<1.
\end{align}
This is equivalent to assuming a lack of confounding between the covariates and the treatment. However the so called ``fundamental problem of causal inference" is that for each observation we only see one of $Y_1(x_i)$ and $Y_0(x_i)$. The one that is not observed is called the {\bf counterfactual outcome} or what would have happened if we had chosen to treat (or not treat) a specific individual. 

Since we are interested in the causal impact of the treatment, two quantities are of interest: the {\em average treatment effect} (ATE) $\Delta=\mathbb{E}(Y_1)-\mathbb{E}(Y_0)$ and the conditioanl average treatmentt affect (CATE)\footnote{Also called the individual average treatment effect in some instances.} $\Delta(x)=\mathbb{E}(Y_1|x)-\mathbb{E}(Y_0|x)$. 
A classic parametric way to estimate ATE is the doubly robust estimator \citep{kang2007demystifying,funk2011doubly, lunceford2004stratification}. The doubly robust estimator can be motivated as follows. Under strong ignorability, we have for any observation $(y_i, t_i, x_i)$
\[
Pr(x_i, t_i, y_i)=Pr(x_i)\underset{\text{treatment model }}{\underbrace{Pr(t_i|x_i)}}\underset{\text{outcome model}}{\underbrace{Pr(y_i|x_i)}}
\]
which suggests fitting separate models for the treatment probabilities $\pi_i=\pi(x_i)=Pr(t_i=1|x_i)$ and $f_1$ and $f_0$ for the outcomes of treated and untreated patients. The values of $\pi_i$ are called the {\em propensity scores}. They are used to do {\em inverse propensity weighting} where the observed population is scaled by $1/\pi_i$ if the person was treated and by $1/(1-\pi_i)$ otherwise. The typical assumption is that $\pi_i, f_1, f_0$ come from parametric families and so can be written $\pi_i(x_i, \beta), f_0(x_i, \alpha_0)$ and $f_1(x_i, \alpha_1)$. These parameters can in turn be estimated giving approximations $\pi_i(x_i,\hat{\beta}), f_0(x_i,\hat{\alpha}_0)$ and $f_1(x_i,\hat{\alpha}_1)$. The doubly robust estimator  uses these to construct two estimators $\Delta_{DR}^j$ of $\mathbb{E}[Y_j]$ for $j\in \{0,1\}$ and estimates the ATE by their difference $\Delta_{DR} = \Delta_{DR}^1 -\Delta_{DR}^0$. It can be shown that
\[
\mathbb{E}(\Delta^1_{DR})=\mathbb{E}(Y_1) + \mathbb{E}\left[
\underset{\text{propensity model error}}{\underbrace{\left(\frac{t-\pi(x_i,\beta)}{\pi(x_i,\beta)}\right)}}
\underset{\text{outcome model error}}{\underbrace{\left(Y_1-f_1(x,\alpha_1)\right)}}
\right],
\]
and similarly for $\Delta^0_{DR}$. Thus if the propensity model or the regression model is correct or very nearly correct then $\Delta_{DR}$ is an unbiased estimate of the ATE $\Delta$. 

The advantage of using $\Delta_{DR}$ is that it allows for accuracy in one model to mitigate the impact of errors in the other. However, the obvious drawback is that both models must be specified. If well-founded parametric families are unavailable then the advantages of double robustness disappear.

Modern approaches in causal inference address this problem by using machine learning techniques to estimate $\Delta$ and $\Delta(x)$. In \citep{athey2015machine,athey2019estimating} a novel tree-based method, called Causal Trees (CT), was proposed. This is based on an extension to typical cross-validation methods used to fit classification trees. For a subset of a partition of the covariate space given by the decision tree, the treatment effect in that subset is given by an inverse propensity weighted estimate similar to $\Delta_{DR}$ for just the covariates in that set. Specifically for all $X_i$ in that partition the causal effect is calculated:
\[
\hat{\Delta}=
\frac{\sum_{i}Y_i *\frac{t_i}{\hat{\pi}(x_i)}}{\sum_i \frac{t_i}{\hat{\pi}(x_i)}} - \frac{\sum_{i}Y_i *\frac{(1-t_i)}{(1-\hat{\pi}(x_i))}}{\sum_i \frac{(1-t_i)}{(1-\hat{\pi}(x_i))}},
\]
where $\hat{\pi}(x_i)$ is an estimated propensity score.

Due to their power as function approximators, neural networks have been a popular recent approach to estimating causal effects. In \citep{johansson16} the same motivation behind inverse propensity weighting ---  reweighting the population such that confounders are equally distributed between the treated and untreated --- is transferred to the representation learning setting. They train a neural network to learn a representation $\Phi(x_i, t_i)$ and a hypothesis (analogous to the outcome model above) such that it minimizes a compound loss
\[
\mathcal{L}(\Phi,h)=\frac{1}{n}\sum_{i=1}^n|h(\Phi(x_i),t_i)-Y_i| + \alpha \text{disc}_{H}(\hat{P}_\Phi^F,\hat{P}_\Phi^{CF})+\frac{\gamma}{n}\sum_{i=1}^n|h(\Phi(x_i),1-t_i)-Y_i|,
\]
where $\text{disc}_H$ is the discrepancy \citep{mansour2009domain} over a hypothesis class $H$, $\hat{P}_\Phi^{F}, \hat{P}_\Phi^{CF}$ are empirically estimated distributions fit over the populations $\{(\Phi(x_i),t_i)\}$ and $\{(\Phi(x_i),1-t_i)\}$, and $\alpha,\gamma$ are positive hyperparameters. Building on this work 
\cite{shalit2017estimating} uses a similar loss function where those related to the hypothesis class are collapsed into a single term with a slightly different weighting and the discrepancy is replaced by an integral probability metric \citep{muller1997integral}. Because there is no explicit closed form for the estimators learned by these models, these neural approaches are evaluated using the \textit{precision in estimation of heterogeneous effect} (PEHE), defined as 
$\epsilon_{PEHE} = \frac{1}{n} \sum_{i=1}^{n}(y_i^{1} - y_i^{0} - \tau(x_i))^2$. 
This notion of error is different from the doubly robust estimators in that it seeks to understand the cumulative error from all observations.
Finally \cite{shi2019adapting} use a neural network with three output channels. Two, written $Q(x_i,1,\theta), Q(x_i,0,\theta)$ are attempts to estimate $\mathbb{E}[Y|X=1]$ and $\mathbb{E}[Y|X=0]$, respectively. The third channel $g(x_i,\theta)$ is meant to model propensities. In all cases $\theta$ represents the neural network parameters. The first loss function is
\[
R(\theta) = \frac{1}{n}\sum_{i=1}^n (Q(t_i, x_i,\theta)-y_i)^2 +\alpha\text{CrossEntropy}(g(x_i,\theta),t_i),
\]
where $\alpha>0$ is again a hyperparameter. The second loss function uses targeted regularization, a technique motivated by connections between causal inference and semiparametric statistics  \citep{kennedy2016semiparametric}. Specifically they create an adjusted prediction $\tilde{Q}(t_i,x_i,\theta,\epsilon) =Q(t_i,x_i,\theta) +\epsilon[\frac{t_i}{g(x_i)}-\frac{1-t_i}{1-g(x_i)}]$, compute residuals $\gamma(t_i,x_i,\theta,\epsilon,y_i)=(\tilde{Q}(t_i,x_i,\theta,\epsilon)-y_i)^2$ and then minimize over $\theta$ and $\epsilon$ the compound loss $R(\theta)+\frac{\beta}{n}\sum_i\gamma(x_i,t_i,y_i,\theta,\epsilon)$. Again $\beta>0,\epsilon\geq0$ are hyperparameters. They evaluate the method by mean absolute error in both in-sample and out-of-sample instances. 

The work above is all in a similar spirit. The propensity and outcome models are decoupled, re-weighted or regularized by propensity, and these transformations are applied to the data in order to approximate the counterfactual distributions. Working in the RCM framework is limited by the fact that the treatment is assumed binary, when many real world questions have multiple treatment options or even a continuum of doses. Further, the assumption of ignorability often does not hold in practice. Structural causal models generalize RCMs and address these issues, but also introduce some novel challenges.

\subsection{Structural Causal Models}\label{ssec:scm}

A line of research in computer science that grew out of approaches in economics \citep{duncan2014introduction,goldberger1973structural} and graphical models \citep{neapolitan2004learning} culminated in the formalization of structural causal models (SCM) \citep{pearl2009causal}. SCMs {\em presume} a set of causal relations derived from expert advice, past experiments, or some other source. SCMs encode these relations as a tuple $\mathcal{M}=(V,U, F, Pa(\cdot),P(U))$ where, $V$ is a set of observed variables, $U$ is a set of unobserved random variables, 
$Pa:S\subset V\cup U\to (V\cup U)-S$ is the {\bf parent function}, $P(U)$ is a probability distribution over the exogenous variables, and $F$ is a family of functions $f_V$ mapping $f_V:Pa(V)\to \text{range}(V)$. 
Intuitively, the SCM puts all the stochasticity into the  $U$ variables and then makes the observed variables $V$ deterministic functions of their parents.

In SCMs the concept corresponding to {\em treatment} is {\em intervention}. Intervening on a particular variable $X$ by setting it to a particular value $\hat{x}$ is written $\text{do}(X=\hat{x})$ and corresponds to replacing $f_X$ with the constant $\hat{x}$ and leaving the rest unchanged.  

To be concrete: one SCM for the example given in the introduction is as follows. Let $V=\{M,A\}$ be the observed variables for a person taking the {\bf M}ulti-vitamin and the {\bf A}pp on their phone. We stipulate that $U_i$ be Bernoulli with parameter $p_i$ for $i\in \{M,A\}$, and that $Y$ is normally distributed with mean $-A$ and unit variance. Intervening by forcing someone to take the multi-vitamin would be written as $\text{do}(M=1)$ and correspond to the modification in \Cref{eq:scm_} below. This has an obvious graphical representation which is given in \Cref{fig:intervention}.
 \begin{align*}
     &\text{base model }\mathcal{M} & &  &\text{intervention model }\mathcal{M}_1& &\\
     A&=U_a & & &A=U_a\\
     M&=\max\{A,U_m\} & \overset{\text{do}(M=1)}{{\huge \Longrightarrow}}& &M=1\numberthis \label{eq:scm_}\\
     Y&=\mathcal{N}(-A,1) & & &Y=\mathcal{N}(-A,1) & & 
 \end{align*}

Clearly the change to $\mathcal{M}_1$ from $\mathcal{M}$, and generally from $\mathcal{M}$ to $\mathcal{M}_x$ under $\text{do}(X=x)$, changes the distribution over the observed variables. Thus if we care about the distribution of $Y$ in $\mathcal{M}_x$ then we are asking about $Pr(Y|\text{do}(X=x))$ which is not necessarily the same as $Pr(Y(|X=x)$. 
The counterfactual quantities of the RCM can be derived from the interventional expressions. $\Delta$ is now equal to the difference in expectations of $Y$ in $M_{t=1}$ and $M_{t=0}$ and the more general conditional average treatment effect is given by the difference in expectation in $Y$ in the corresponding to the interventions $\text{do}(X=x_i,t=1)$ and $\text{do}(X=x_i,t=0)$. 

To craft estimators of these quantities (and others) in the SCM framework one must use the do-calculus \citep{pearl1995causal}, which is a means of rewriting {\em causal queries} of the form $Pr(Y|\text{do}(X=x))$ such that they only contain values derived from the {\em observational distribution} $Pr(V)$.\footnote{We are glossing over technical issues about when the do-calculus can be guaranteed to be successful} The do-calculus also contains a specific set of graphical criteria for knowing which variables need to be measured to ensure conditional independence/unconfoundedness \citep{pearl2009causal}.

Thus constructing estimators for causal effects in the SCM framework is relatively simple.  Given a SCM $\mathcal{M}$, a data set of the observed variables $V$, and a causal query, one first estimates the joint distribution $P(V)$, then rewrites the causal query using the do calculus, and then uses the estimated joint to approximate the quantities derived from the do-calculus.

The advantages of the SCM approach are clear: counterfactuals are a derived rather than assumed quantity, treatments can be arbitrary, and ignorability can be explicitly checked. However, the drawbacks are that the estimators are dependent on having preconstructed causal model that may be incorrect and which cannot be explictly checked for correctness against a ground truth. For cases where the SCM is well-founded or simple (such as MDPs), however, the SCM framework provides a powerful inference engine.

Having reviewed two causal frameworks we give some examples of their application to sequential decision making problems.

\section{Causality in Decision Making: A Bandit Warmup}\label{sec:bandits}
We choose multi-armed bandit problems\citep{lattimore2018bandit} as a natural bridge MDP-based reinforcement learning. Multi-armed bandit problem proceed in rounds. At each round the agent chooses an action  $a_t$ in a finite set $\mathcal{A}$ and receives a reward for that action $r_t$. In this case a policy reduces to picking the arm with the largest expected reward. A generalization is the {\em contextual} bandit problem were at round $t$ a context $x_t$ in a context set $\mathcal{X}$ and a $|\mathcal{A}|$-dimensional reward vector $r_t$ are drawn from a distribution $\mathcal{D}$. The agent chooses an action according to a policy $\pi(x_t)$ and receives the component of $r_t$ corresponding to the action $\pi(x_t)$.

We explore causal connections only briefly as full treatments are beyond the scope of this project. However, they provide a stepping stone to causality in the full RL setting and make clear the correspondence between causal inference and off-policy evaluation in batch RL, on which we shall elaborate in \cref{sec:OPPE}.

\paragraph{RCMs in Contextual Bandits}

Connections between RCMs and contextual bandits arise in the setting of {\em off-policy evaluation} or learning from logged bandit feedback \citep{swaminathan2015counterfactual}. In this case a fixed data set $S$ of actions, contexts, and rewards drawn from $\mathcal{D}$ and some {\em logging policy} is given to the agent where the actions were chosen by one or more unknown policies. There are two related but distinct goals in this setting: policy {\em evaluation} which tries to use $S$ to estimate the value of some policy $\mu$ and policy {\em optimization} which aims to learn an optimal policy from $S$. The seminal contributions in this field address both of these questions \citep{dudik2011doubly, dudik2014doubly}. In \citep{dudik2011doubly} the value of a policy is estimated using a doubly robust estimator similar to those discussed in \Cref{sec:caus}. Here the estimate is unbiased if the policy or the reward distribution are accurately approximated. 

Another approach uses the counterfactual risk minimization principle \citep{swaminathan2015counterfactual} to devise off-policy estimators in a principled way that also uses propensity weightings. Follow-on work addressed the potential for overfitting in the propensities  \citep{swaminathan2015self}. 

\paragraph{SCMs in Bandit Problems}

Combining SCMs and (contextual) bandit problems usually takes the form of imposing an SCM structure on the problem. \cite{lattimore2016causal} designate a binary reward variable $Y$, assume an SCM structure over $Y$ and the context variables, and define the action space to be interventions of the form $a_t=\text{do}(X_t=x)$. The agent picks an intervention, then for the remaining excluding $X_t$ (written $X^C$) they observe $P(X^C|\text{do}(X=x)$. Regret bounds are given for specific graph topologies as well as general graphs. 

In \citep{bareinboimbandits15}, the authors provide an SCM for the typical bandit decision making problem (Figure 2 in the referenced paper) and show that if that SCM has unobserved confounders then typical bandit algorithms are suboptimal and present a variant on Thompson Sampling for this scenario. In \citep{lee2018structural,lee2019structural} they give a variant called an SCM-MAB with the goal of trying to estimate a possibly empty set of interventions to alter the behavior the underlying system and extend to the case where some interventions are not allowed.

\section{Batch Reinforcement Learning and Off-Policy Evaluation}\label{sec:batchRL}

Different from contextual bandit problems, where contexts do not depend on past actions, RL studies sequential decision-making in unknown environments, where states are related to past actions. As we shall see more clearly in \cref{sec:OPPE}, RL is closely related to bandit problems and can be ``decomposed'' into contextual bandits. Generally, in a RL problem, the task of the agent is to find an optimal policy maximizing the expected long term return, based on the rewards received at each time step. The challenge here is that the agent does not have complete information about the dynamic environment and tries to identify high-reward behavior patterns by interacting with the unknown environment. If interactions are not allowed and only historical experiences are available, then the agent has to learn a best possible policy from the fixed dataset, which is referred to as offline or batch reinforcement learning.

In this section, we first give a quick review on Markov Decision Process (MDP) \citep{Bel}, which is the mathematical formulation of reinforcement learning. Then, we give an overview of batch reinforcement learning algorithms and we argue that the core of batch reinforcement learning is actually policy evaluation problem. At the end of this section, we provide the formulation of off-policy evaluation problem and briefly introduce non-causality based evaluation techniques, with which we shall compare those causal approaches.    

\subsection{Markov Decision Process}
A Markov Decision Process \citep{sutton2018reinforcement} is denoted by a tuple $\MDP$, where $\sset$ is the state space, $\aset$ is the action space and $\rset$ is the set of rewards. At a given time step $\step$, the agent takes an action $a_t\in \aset$ in the state $s_t\in \sset$ and receives a reward $r_t$ that is determined by the \textit{reward function} $\Rfun: \sset\times\rset\times\sset\rightarrow \rset$, i.e., $r_t:=\Rfun(s_t,a_t,s_{t+1})$, where $s_{t+1}$ denotes the next state following the distribution in accordance to the \textit{transition dynamics} $\Tdef$, $\Tfun(s_t,a_t,s_{t+1})\coloneqq\Pr(s_{t+1} \mid s_t, a_t)$. Finally, $\isset$ is the initial distribution of states defined as $\issetdef$ and $\D$ is the discount factor defined as $\Ddef$.  

As we mentioned above, the goal of the agent is to find a policy $\pdef$ defined as a probability distribution over $\sset\times\aset$, such that by following this policy, the agent maximize the expected return. For the sake of simplicity, we here consider finite horizon MDP, where trajectories $\tau=\left(s_{0}, a_{0}, \dots, s_{H}\right)$  are sequences of states and actions of finite length up to $s_H$, the terminal state. We note that if a trajectory $\tau$ is generated by following a policy $\mu$, then $p_\mu$, the distribution of $\tau$, can be explicitly computed as follows, given the transition probability $\Tfun(s,a,s')$: $p_\mu=d_{0}\left(s_{0}\right) \prod_{t=0}^{H-1}\left[\Tfun(s_t,a_t,s_{t+1}) \mu\left(s_{t}, a_{t}\right)\right]$, where $d_0$ is the initial state distribution. Therefore, the ultimate goal for the agent is to find a policy $\pi$ such that the expected return of this policy $\mathbb{E}_{\tau\sim p_\pi}\{\sum_{i=0}^{H-1} r_t\}$ is maximized. Generally we treat the expected return as the quality of the policy and as we shall see later, the estimation of the policy quality is of vital importance in RL. Besides this expected return, another way to evaluate a policy is to look at the conditional expected return that is conditional on the initial state $s_0$, which is referred to as the value function $V^{\pi}(s):=\mathbb{E}_{p_\pi}\{\sum_{i=0}^{H-1} r_t|s_0=s\}$. Similarly, the Q-function $Q^\pi(s,a):=\mathbb{E}_{p_\pi}\{\sum_{i=0}^{H-1} r_t|s_0=s, a_0=a\}$, the expected return conditional on both the initial state $s_0$ and the initial action $a_0$, also measures the performance of the given policy $\pi$.

\subsection{Batch Reinforcement Learning}
In principle, standard off-policy reinforcement learning algorithms such as deep Q-learning (DQN)\citep{mnih2015DQN} and deep deterministic policy gradient \citep{lillicrap2015continuous} are applicable in the batch setting, since they also aims at learning the optimal policy based on the samples generated by a different behavioral policy. However, as observed in a prior work on off-policy learning \citep{fujimoto2019off}, some off-policy deep reinforcement learning algorithms fail in the batch setting because evaluating state-action pairs (more generally, policies) that are not contained in the provided batch of data is very challenging and estimators learned from the fixed dataset generalize poorly in this setting and is unable to provide high-quality estimates due to extrapolation error. In other words, policy evaluation based on off-policy data is the crux of batch reinforcement learning. 

In order to better evaluate state-action pairs or polices in batch reinforcement learning, many algorithms have been proposed, which fall mainly in two categories. The first relies on policy constraints, which constrains the learned policy to a subset of policies which can be adequately evaluated, rather than the process of evaluation itself \citep{fujimoto2019off, laroche2019safe, kumar2019stabilizing}. More intuitively, these algorithms require that the learned policy cannot deviate too much from the behavioral policy so that the estimators based on prior experiences can return fairly good estimates. For more details, we refer reader to the survey by \cite{fujimoto2019benchmarking}.

Here we mainly address the other category that aims at improving the estimators such as Q functions with various techniques for reducing extrapolation error. Quantile Regression DQN (QR-DQN) proposed by \cite{dabney2018distributional} leverages a distributional reinforcement learning method \citep{morimura2010nonparametric} for a better Q function. Simply put, QR-DQN tries to reduce the error by first computing quantiles of the return distribution and then taking the average of these quantiles as the estimation. Similarly, Random Ensemble Mixture (REM) DQN \citep{agarwal2019striving} also maintains the estimates of multiple Q functions which are all trained with different target networks and finally a convex combination of these estimates gives an improved Q value estimation.     

Evidently, estimator construction or more essentially policy evaluation using off-policy data lies at the heart of batch reinforcement learning and current evaluation techniques or estimators often suffer from poor generalization (extrapolation error). Recent advances in off-policy evaluation indicate that causal inference is very helpful when constructing estimators that generalize well to those actions not contained in prior experiences, as we shall see in the following section. Before moving to causality-based policy evaluation, we first give the problem formulation for off-policy evaluation and an overview of existing methodologies that are not causality-based, which will be compared with causal approaches, showing the advantages of incorporating causality.    

\subsection{Off-Policy Evaluation}
Policy evaluation is about measuring the quality of a given policy, and this measurement rests on the sample trajectories either generated by the policy to be evaluated (on-policy) or by some different policy or policies (off-policy). Generally, off-policy policy evaluation is more challenging than on-policy one, since there is a distributional mismatch between the policy we are interested in and the policy  
that is actually implemented in the MDP model for generating those sample trajectories. In the sequel, we refer to the interested policy as the target policy, denoted by $\pi$ and the data-generating policy as the behavioral policy $\mu$. To see this mismatch, we only need to realize that sample trajectories are in fact a series of state-action pairs sampled from the trajectory distribution related to the behavioral policy, i.e., $p_{\mu}$ and for the target policy $\pi\neq\mu$, we have the mismatch $p_{\mu}\neq p_{\pi}$, meaning that this sample trajectory $\tau$ cannot be directly applied to estimating the expected return $\mathbb{E}_{\tau\sim p_{\pi}}\{\sum_{i=0}^H r(s_t,a_t)\}$. In order to address this distributional mismatch, many approaches have been proposed, we categorize these methods into three groups: \textbf{model-based}, \textbf{model-free}, and \textbf{hybrid} ones. Here, we merely give a bird's eye view of existing off-policy evaluation techniques and detailed discussions are included in the next section, where we look into the difference between them and those causality-based  off-policy policy evaluation (OPPE) approaches. 
\subsection{Model-based Off-Policy Evaluation}
These methods first learn an MDP model from the observations (sample trajectories) and then use the learned model to evaluate the target policy. Specifically, model-based methods focus on constructing an approximation of the reward function $r(s,a)$ and/or the transition kernel $\Tfun(s,a,s')$. For example, approximate models in prior works \citep{jiang2016doubly, thomas2016data} rely on a reward function estimate of the underlying MDP model. For model estimation, the simplest method is probably the count-based estimation for discrete MDPs, where the reward function and the transition kernel are approximated based on observed state-action-reward tuples, leading to consistent maximum-likelihood estimators of the model. However, this method has no or limited generalization \citep{paduraru2013off}. If no samples are available for some state-action pair, then the associated estimation is not defined. 

Equipped with  generalization ability, another common approach is regression estimator \citep{paduraru2013off, mannor2007bias} for continuous cases, which is essentially a representation learning for the MDP and its performance heavily depends on the selection of representation class and the design of loss functions. However, as pointed out by \cite{farajtabar2018more}, there are two major problems with these model-based approaches: (1) Its bias cannot be easily quantified, since in general it is difficult to quantify the approximation error of a function class, and (2) It is not clear how to choose the loss function for model learning. The first problem is more about a choice of representations and falls within the realm of function approximation problem in RL\citep{li2017deep,li2019convergence, geramifard13linear}. For the second one, causal inference can bring a more reasonable loss function for model learning and we shall compare model-based estimators with causality-based model learning at the end of the next section, where we show the advantage of counterfactual thinking during the model learning process.  
\subsection{Model-free Off-Policy Evaluation}
Model-free approaches do not rely on the underlying MDP model. Most model-free methods \citep{guo2017using, precup2000eligibility, thomas2015high} aim directly at the distributional mismatch by importance sampling \citep{kahn1953methods}. A technique from computational statistics, Importance Sampling (IS) computes an expected value under some distribution of interest by weighting the samples generated from some other distribution, which exactly fits the OPPE formulation. As noted above, a sample trajectory $\tau=\left(s_{0}, a_{0}, \dots, s_{H}\right)$ with a distribution $p_{\mu}$, the empirical return $\sum_{t=0}^{H-1}r_t$ gives an unbiased estimate of the expected return $\mathbb{E}_{\tau\sim p_{\mu}}\{\sum_{i=0}^H r(s_t,a_t)\}$. If we define the per-step importance ratio as $\rho_t:=\pi(s_t,a_t)/\mu(s_t,a_t)$ and the cumulative importance ratio as $\rho_{0:t}:=\prod_{k=0}^t \rho_k$, then the basic (trajectory-wise) IS estimator, and an improved step-wise version are given as follows:    
$V_{\mathrm{IS}}:=\rho_{1: H} \cdot\left(\sum_{t=1}^{H} \gamma^{t-1} r_{t}\right),V_{\mathrm{step}-\mathrm{IS}}:=\sum_{t=1}^{H} \gamma^{t-1} \rho_{1: t} r_{t},$ which is based on the fact that $p_{\pi}=d_{0}\left(s_{0}\right) \prod_{t=0}^{H-1}\left[\Tfun(s_t,a_t,s_{t+1}) \mu\left(s_{t}, a_{t}\right) \rho_t\right]$. For more details and variants such as weighted importance sampling, see \cite{precup2000eligibility, thomas2015high}. The advantage of model-free methods over model-based ones is obvious. Model-free methods provide a consistent estimator with low bias, whereas model-based ones often suffer from higher bias, if the estimated model is a poor approximation. 

An IS-based approach has two major issues. First, this approach tends to have high variance, especially when the target policy is deterministic. Second, it requires absolute continuity assumption regarding the polices. That is, for all state-action pairs $(s,a)\in \sset\times\aset$, if $\mu(s,a)=0$ then $\pi(s,a)=0$. This assumption requires that all state-action pairs $(s,a)$ produced by the target policy must have been observed in the sample trajectories, otherwise the importance ration $\rho_t$ cannot be defined. To sum up, there is no generalization in an IS approach, meaning that we are unable to deal with the data we haven't seen before. In the following we shall see how causality-based model-free methods tackle these issues with counterfactual inference, enabling generalization in off-policy evaluation when facing unobserved data.   
\subsection{Hybrid Off-Policy Evaluation}	
There has been a growing interest in combining the two approaches for constructing unbiased estimators with low variances. Building off the  named \textit{doubly robust} (DR) estimator \citep{dudik2011doubly} discussed in \cref{sec:bandits}, \citep{jiang2016doubly} developed this technique for off-policy evaluation in reinforcement learning.  By decomposing the whole trajectory into state-action-reward tuples, off-policy evaluation in reinforcement learning can be viewed as estimation in multiple contextual bandits problems, where $s_t$ is the state, $a_t$ is the action, and the observed return is defined recursively for adding the temporal relations in MDP to the rewards in the contextual bandit setting. Built upon this DR estimator, many variants have been proposed recently. For example,  \cite{farajtabar2018more} propose a loss function (a weighted mean square error) in model learning for reducing the variance, yielding a better model estimation, which minimize the variance of the DR estimator. On the other hand, a novel approach for blending model-based and model-free parts is considered by \cite{thomas2016data}. In this paper, model-based and model-free estimators are not applied to the sample trajectory at the same time, instead, they first partition the whole trajectory into to parts, where the first half is handled by importance sampling and the rest by a model-based estimator. Naturally, different partitions lead to different estimates, blending model-free and model-based estimates together differently and the new estimator in this paper returns a weighted average of these blending estimates so as to minimize the mean square error.            

To sum up, we find that current approaches often suffer from poor generalization, leading to poor estimates in off-policy evaluation. However, if we take a new angle that is different from the statistical or MDP-based ones, we can see that generalization is actually equivalent to the measurement of counterfactual outcomes, state-action pairs that have never happened before. From this perspective, causal inference is of great help when dealing with off-policy data, since it reveals the causality behind data that can be used for producing better generalization, as we shall present in the next section.     
\section{Off-Policy Policy Evaluation: Causal Approaches}\label{sec:OPPE}

The idea behind causality-based off-policy policy evaluation is that the target policy is treated as a kind of intervention, comprising counterfactual actions that are different from the those in behavioral policy. Therefore, in order to evaluate the target policy based on observational data generated by the behavioral policy, we have to focus on the difference between the two policies or more specifically, the counterfactual outcomes. That is, \textit{what would have happened, had we applied those actions from the target policy}. In other words, with this counterfactual thinking, OPPE problem reduces to counterfactual inference problem and in this subsection, we shall present two approaches of leveraging causal inference in solving OPPE problem.   
\subsection{Model-free Approach: Counterfactually-guided Policy Evaluation}
The first approach does not rely on the underlying \textbf{MDP models}, and on the contrary, it casts MDPs into structrual causal models (SCM) for counterfactual inference, where actions in the target policy are viewed as interventions. We first introduce the generic approach of counterfactual inference then we move to its applications in OPPE, as studied by \cite{buesing2018woulda,  oberst2019counterfactual}. 

As we have demonstrated in \cref{sec:caus}, counterfactual outcomes can be measured by counterfactual inference technqiues, which estimate the answer to the counterfactual question \textit{what would have happened, had we applied an intervention}.

On the other hand, as we discussed before, OPPE is also about answering a counterfactual question \textit{what would have happened, had we applied those actions from the target policy}. Hence, if we can represent any given MDP models or more broadly POMDP models by an SCM over trajectories \citep{buesing2018woulda}, and we treat those trajectories generated by the behavioral policy as observations, actions from the target policy as interventions, then we are also able to identify the expected return under the target policy using counterfactual inference, where the value function becomes the query variable. This one-to-one correspondence between counterfactual inference and OPPE, as shown in Table\ref{cfi_oppe}, is built upon structural models that capture the underlying cause-effect relationship between variables, which has been the foundation for some research works on causality-based decision-making, such as multi-armed bandits problem \citep{bareinboimbandits15, zhang2017transfer} and reinforcement learning \citep{buesing2018woulda, oberst2019counterfactual}. In the sequel, we detail the SCM for addressing OPPE and explicitly express evaluation of the target policy as an intervention.
\begin{table}
\small
\centering
	\begin{tabular}{ll}
	\toprule
	CFI& OPPE\\
	\midrule
	observations & off-policy episodes\\
	interventions  & target policy\\
	query variable  & expected return \\
	\bottomrule
\end{tabular}
\caption{Correspondence between Counterfactual Inference (CFI) and Off-Policy Policy Evaluation (OPPE)}
\label{cfi_oppe}
\end{table} 

\begin{figure}[ht]
	\centering
	\includegraphics[width=0.6\textwidth]{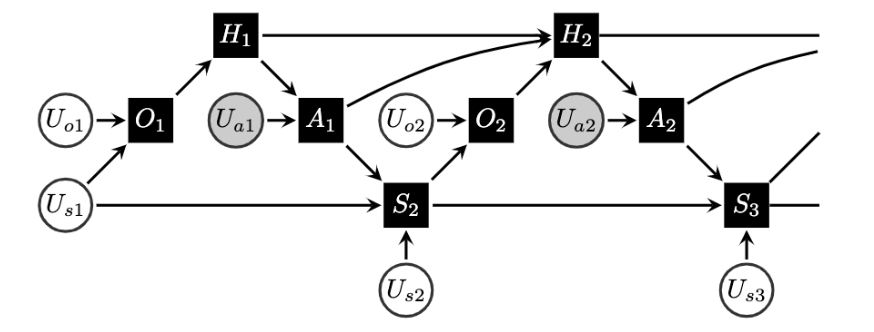}
	\caption{DAG for the POMDP example, taken from \citep{buesing2018woulda}: $\st{t}$ denotes the state at time $t$. $H_t:=(O_1,A_1, O_2,A_2, \cdots, A_{t-1}, O_{t})$ is the set of historical observations up to time $t$. $U_{o_t}, U_{s_t}$ denote the unobserved confounders for the observation $O_t$ and $\st{t}$ respectively.}
	\label{dag_pomdp}
\end{figure}

In MDP (POMDP) or bandits problems, the casual relationships among variables are straightforward and we only need to care about those unobserved confounders when constructing SCMs. For example, in POMDP, the agent makes a decision based on the history he has observed, which consists of the observation at each time step and the observation is determined by the state the agent is currently in. Meanwhile, the decision taken at this time step will lead the agent to the next state, which in turn influence future's observations. The DAG for this POMDP example is provided in Fig\ref{dag_pomdp}, where the unobserved confounders in POMDP are also included. According to the DAG, in the SCM, we can express the transition kernel $\Tfun(\st{t},\at{t},\st{t+1})$ as deterministic functions with independent noise variables $U$, such as $\st{t+1}=f_{st}(\st{t},\at{t},U_{st})$, which is always possible using auto-regressive uniformization \citep{buesing2018woulda}. Accordingly, we express a given policy $\pi$ as a causal mechanism: $\at{t}=f_\pi(H_t,U_{at})$, therefore, running the target policy $\pi$ instead of the behavioral policy $\mu$ in the environment can be viewed as an intervention $I(\mu\rightarrow \pi)$ consisting of replacing $\at{t}=f_\mu(H_t,U_{at})$ by $\at{t}=f_\pi(H_t,U_{at}).$ We summarize the model-free approach in Algorithm \ref{cfi_oppe_algo}.

\begin{algorithm}

\small
\caption{Counterfactual Policy Evaluation}
\begin{algorithmic}[1] 
	\Procedure{CFI}{observations $\hat{x}_0$, SCM $\mathcal{M}$, intervention $I$, query variable $X_q$} 
	
	\Comment{This procedure performs counterfactual inference based on the SCM and observations}
	\State $p(u|\hat{x}_0)\leftarrow (\mathcal{M}, \hat{x}_0)$ \Comment{compute the posterior based on the observations and the SCM}
	\State $\hat{u}\sim p(u|\hat{x}_0)$ \Comment{sample noise from the posterior}
	\State $p(u)\leftarrow \delta(u-\hat{u})$ \Comment use the sample distribution as the noise distribution in SCM
	\State $f\leftarrow f^{I}$ \Comment perform the intervention and change the causal mechanisms accordingly
	\State \Return $x_q\sim p^{\mathrm{do}(I)}\left(x_{q} | \hat{u}\right) $
	\EndProcedure

	\Procedure{CFI-OPPE}{SCM $\mathcal{M}$, target policy $\pi$, historical dataset $D$, number of samples $N$} 
	
	\Comment{This procedure performs OPPE based on CFI} 
	\For{$i\in\{1,2,\cdots, N\}$}
	\State $\hat{h}_{T}^{i} \sim D$ \Comment sample from the dataset
	\State $g_{i}=\operatorname{CFI}\left(h_{T}^{i}, \mathcal{M}, I(\mu \rightarrow \pi), G\right)$ \Comment Counterfactual Inference 
	\EndFor
	\State \Return $\frac{1}{N} \sum_{i=1}^{N} g_{i}$  
	\EndProcedure
\end{algorithmic}	
\label{cfi_oppe_algo}
\end{algorithm}
   
\subsection{Model-based: Counterfactually-guided Model Learning}
Different from model-free methods that completely bypass MDP models, model-based ones still rely on the underlying MDP model learned from the observational data generated from the behavioral policy for evaluating the target policy. One of the key challenges in model leaning is the choice of the loss function for model learning without the knowledge of the evaluation policy (or the distribution of the evaluation policies). Without this knowledge, we may select a loss function that focuses on learning the areas that are irrelevant for the evaluation policy. Different from previous model-based methods which only take into account data generated by behavioral policy, a new loss function is proposed involving counterfactual thinking when learning the representation using neural networks \citep{liu2018representation}, which includes target policy evaluation in the representation learning process. This makes the learned model more suitable for OPPE, compared with models only involving behavioral data.    

To be more specific, for an unknown MDP model $M=\langle r(s,a), \Tfun({s,a,s'})\rangle$, we can learn a model $\widehat{M}=\left\langle\widehat{r}(s, a), \widehat{\Tfun}\left(s, a, s'\right)\right\rangle=\left\langle h_{r}(\phi(s), a), h_{\Tfun}\left(\phi(s\right), a,\phi(s^{\prime})\right\rangle$, which is based on the representation function $\phi$, e.g., a neural network. The procedure of model-based approach comprises two steps: we first learn the representation $\phi$ based on the historical data and a carefully selected loss function, derived from counterfactual thinking. Then, we evaluate the target policy using the learned model $\widehat{M}$, which is the same as the classical model-based methods. In this subsection, we mainly focus on the first step, presenting how counterfactual thinking leads to a suitable representation for evaluating the target policy in model learning. Before we move to the details, we first introduce the following notation.  For $\tau=\left(s_{0}, a_{0}, \dots, s_{H}\right)$, a trajectory from model $M$, generated by the policy $\mu$, we define the joint distribution of $\tau$ as $p_{M,\mu}=d_{0}\left(s_{0}\right) \prod_{t=0}^{H-1}\left[\Tfun(s_t,a_t,s_{t+1}) \mu\left(s_{t}, a_{t}\right)\right]$. Accordingly, we can define the associated marginal and conditional distributions as $p_{M, \mu}\left(s_{0}\right), p_{M, \mu}\left(s_{0}, a_{0}\right)$. With these distributions, we can define the $t-$step value function of policy $\mu$ as $V_{M,t}^\mu(s)=\mathbb{E}_{\tau\sim p_{M,\mu}(s_0)}\{\sum_{i=0}^t r_i\}$. Similarly, for the estimated model $\widehat{M}$ and the target policy $\pi$, we can define distributions and value functions in the same way.

Now we are in a position for illustrating the combination of counterfactual thinking and mode-based OPPE that leads to a more data-efficient framework for evaluating the target policy, which is first introduced by \cite{liu2018representation}. Specifically, the key of this framework is that this combination provides a loss function that takes target policy into consideration by counterfactual inference, resulting in a learned model that is suitable for evaluating the tagret policy. Theoretically, for the behavioral policy $\mu$ and the target policy $\pi$, the actual difference is measured by $\mathbb{E}_{s_0\sim \isset}\left(V_{M}^{\pi}\left(s_{0}\right)-V_{M}^{\mu}\left(s_{0}\right)\right)^2$. However, we do not know the underlying MDP model $M$ and can only estimate it based on history data, which we shall detail in the sequel. For now, we assume a learned model $\widehat{M}$ is available, then OPPE relies on the estimated difference$(V^{\pi}_{\widehat{M}}\left(s_{0}\right)-V^{\mu}_{\widehat{M}}\left(s_{0}\right))$. One natural question is how different are the actual difference $\left(V_{M}^{\pi}\left(s_{0}\right)-V_{M}^{\mu}\left(s_{0}\right)\right)$ and the estimated one $(V^{\pi}_{\widehat{M}}\left(s_{0}\right)-V^{\mu}_{\widehat{M}}\left(s_{0}\right))$? To answer this question, we first look at the following upper-bound, which is straightforward by using Cauchy-Schwartz inequality. $\frac{1}{2} \mathbb{E}_{s_{0}\sim\isset}\left[(V_{\widehat{M}}^{\pi}(s_{0})-V_{\widehat{M}}^{\mu}\left(s_{0})\right)-(V_{M}^{\pi}(s_{0})-V_{M}^{\mu}(s_{0}))\right]^{2} \leq \mathrm{MSE}_{\pi}+\mathrm{MSE}_{\mu},$
  where $\text{MSE}_\pi:=\mathbb{E}_{s_{0}\sim \isset}\left(V_{\widehat{M}}^{\pi}\left(s_{0}\right)-V_{M}^{\pi}\left(s_{0}\right)\right)^{2}$ and $\text{MSE}_\mu:=\mathbb{E}_{s_{0}\sim\isset}\left(V_{\widehat{M}}^{\mu}\left(s_{0}\right)-V_{M}^{\mu}\left(s_{0}\right)\right)^{2}$ are the mean-squared error for two policy value estimates. The above inequality tells that the difference is upper-bounded by the sum of two MSEs, and once we can find a representation $\phi$ such that this upper-bound $\text{MSE}_\pi+\text{MSE}_\mu$ is minimized with respect to the representation, then the estimated difference serves as a fairly good evaluation of the target policy. Therefore, the problem reduces to expressing the upper-bound as a functional of $\phi$, since $\widehat{M}$ is merely a notation and it is the reward function $\hat{r}(s,a)=h_{r}(\phi(s), a)$and the transition kernel $\hat{P}=h_{\Tfun}\left(\phi(s\right), a,\phi(s^{\prime})$ that explicitly depend on the $\phi$. This is where counterfactual inference plays a part, as we shall see in the following.
  
  For $\text{MSE}_\mu$, we know the behavioral policy and have the sampled trajectories, hence for this on-policy case, we can apply the Simulation Lemma \citep{kearns2002near}, showing that this on-policy mean square error can be upper bounded by a function of the reward and transition prediction losses. However, the difficulty here is that we do not have on-policy data for the target policy $\text{MSE}_\pi$ and this is why we need counterfactual thinking for constructing an upper bound for $\text{MSE}_\pi$. Specifically, the value function error can be decomposed \citep{liu2018representation} into a one-step reward loss, a transition loss and a next step value loss, with respect to the on-policy distribution. Then, this becomes a contextual bandit problem, and estimation of the MSE is equivalent to the estimation of treatment effect, where the intervention is replacing the behavioral policy $\mu$ with the target policy $\pi$. This approach is built upon the work of \cite{shalit2017estimating} about binary action bandits, where the distribution mismatch is bounded by a representation distance penalty term. Here we merely provide the high-level analysis for estimating $\text{MSE}_\pi$ using causal inference, for more details, we refer readers to research works by \citep{liu2018representation, shalit2017estimating, johansson16}. We consider the following $H-t$ step value error, which decompose the MSE stage-wise, 
  $\epsilon_{V}(\widehat{M}, H-t):=\int_{\sset} \left(V_{\widehat{M}, H-t}^{\pi}(s)-V_{M, H-t}^{\pi}(s)\right)^{2} p_{M, \mu}\left(s_{t} | a_{0: t-1}=\pi\right) d s_{t}.$ From this definition, it is easy to see that $\epsilon_{V}(\widehat{M}, H)=\text{MSE}_\pi$ and more importantly, the difference between $\epsilon_{V}(\widehat{M}, H-t)$ and $\epsilon_{V}(\widehat{M}, H-t-1)$ is related to a treatment effect estimation in a contextual bandit problem, where the state-action-reward tuple is generated from the sample trajectory in reinforcement learning. In this special contextual bandit, the treatment effect estimation measures the counterfactual outcomes of running the target policy. Finally, as we estimate $\text{MSE}_\pi=\epsilon_{V}(\widehat{M}, H)$ using the treatment effect estimation recursively, we combine the upper-bounds for $\text{MSE}_\mu$ and $\text{MSE}_\pi$ and make the combination the loss function for model learning, which leads to a MDP model that is suitable for off-policy evaluation.    

\subsection{Analysis of Causality-based Off-Policy Evaluation}   
If we dive deeper into these causal approaches, we can find that it is the generalization brought up by causal inference that helps solve OPPE problems. Intrinsically, dealing with off-policy evaluation is nothing else more than making a prediction about the decision-making system under counterfactual interventions. From this perspective, we argue that causality enhance the generalization of previous off-policy evaluation methods by transferring the cause-effect relationship we infer from observations to the target policy evaluation process.  Specifically, in this subsection, we compare causality-based approaches with existing off-policy evaluation techniques as we briefly introduced in \cref{sec:batchRL}, showing that how causality equips these novel approaches with better generalization. 

Traditional model-free methods, as we have briefly reviewed above, are rather restrictive in the sense that they often need absolute continuity assumption for computing the importance ratio $\rho_t=\pi(s_t,a_t)/\mu(s_t,a_t)$ and besides this, as we can see from the definition of $\rho_t$, this ratio requires the explicit knowledge of the behavioral policy, i.e., the distribution regarding $\mu$. However, in practice, it so happens that we either do not know the behavioral policy $\mu$ or the sample trajectories are in fact generated by multiple behavioral policies. Moreover, absolute continuity is usually not satisfied in real-world applications, which is a major limit of its wider applications, since there is no generalization within this model-free framework. Even though, as mentioned in prior works \citep{jiang2016doubly, farajtabar2018more}, when the importance ratio is not defined or the behavioral policy is unknown, they can be estimated from the data using some parametric function classes, we think that these remedies do not address the poor generalization of importance sampling directly, which heavily depends on the quality of observations, since they merely try to create the missing pieces of importance sampling.

Causality makes it possible that estimators that do not rest on MDP models are still able to generalize well by incorporating SCMs. Though one may argue that this is not model-free anymore, as this approach involves SCMs, we believe that requiring causal models is the price we have to pay for bypassing the MDP models while acquiring generalization ability, which we shall illustrate later. We first clarify that the term ``model-free'' indicates that this causal approach circumvents complicated MDP models involving both reward and transition kernel estimation and turns to causal models, which are much easier to deal with, since causal relationships are straightforward in MDP setting. Now we are in a position to elaborate on the advantages of this causality-based model-free approach, as shown below.

  \paragraph{Generalization} As we have introduced above, this causal approach does not address the distribution mismatch by importance sampling, instead it first provides a SCM, based on which a counterfactual inference is conducted to evaluating the target policy. In other words, as long as the historical observations produce a good estimate of the unobserved confounders and causal mechanisms, we do not care whether absolute continuity is satisfied or not, since it is no longer the key of the estimation. This generalization from observed actions in sample trajectories to counterfactual actions from the target policy is essentially based on the estimated SCM . 
  \paragraph{Unbiased Estimation} Another advantage of this causal approach is it gives unbiased estimates, if the trajectory distribution of the behavioral policy entailed by the SCM coincides with the actual one distribution entailed by the MDP. This shows that causality-based approach returns a consistent evaluation of the target policy, which is different from MDP model-based estimation.  

On the other hand, different from traditional model-free methods, most current model-based methods comprising model learning/estimation process inherently possess generalization to some extent, since the learned model parameters can be directly leveraged when dealing with a new policy without any alterations. However, without causality, we are unable to acquire some prior knowledge about the target policy and generalize from observations, since the common way for designing the loss function is to look at the empirical risk of evaluating the behavioral policy using the model estimation, which reduces the problem to the on-policy model learning. As argued by \cite{liu2018representation}, this kind of model learning yields a model that is suitable for evaluating the behavioral policy instead of the target policy, since the representations are chosen in such a way that the estimation error of the behavioral policy is minimized. Therefore, in order to learn a MDP model that is suitable for off-policy evaluation, the target policy must be taken into account when designing the model learning, e.g., the loss function. To be more specific, this causality-based approach measures the mean square error of the target policy by investigating the treatment effect as introduced in the second section. In other words, the difference between one treatment, i.e, the target policy and another treatment, i.e., the behavioral policy is measured by causal inference and is further leveraged to design the loss function, as the treatment effect upper bounds the mean square error. To wrap up, causality enhances the generalization of model-based approaches in the sense that it helps the model generalize in a desired direction, producing a more accurate off-policy evaluation.

\section{Other comparisons}
We close by offering two final areas of evidence that reinforcement learning and causality tackle the same problems with similar approaches.

\subsection{Exploring data}

Causality and reinforcement learning have similar approaches in discovering new information about the problem. In both domains, continuing to sample variables increases knowledge of the domain space and decreases expected error in the result.

\subsubsection{Exploring the domain space in RL}
The idea of better exploring a problem domain is made explicit by \cite{hanna2017data}, who introduce  the notion of a \textit{behavior policy}, a policy for exploring the RL problem domain which will minimize the offline estimates for batch reinforcement learning. They define an equation 

$$\frac{\partial}{\partial\theta}MSE[IS(H, \theta)]= \mathbb{E}[-IS(H, \theta)^{2} \sum_{t=0}^{L} \frac{\partial}{\partial\theta}log \pi_{\theta}(A_t|S_t)],$$
where $IS(H, \theta)$ is the weighted importance sampling return of a trajectory, $H$, sampled from a behavior policy, $\pi_\theta$ with parameters $\theta$, minimizing the mean squared error in the reward estimate. The authors propose an update algorithm based on the above gradient to continually refine the behavior policy so as to reduce the error estimate. The prescription for what to sample next, $\pi_\theta$, is thus updated after every sampling.

\subsubsection{Exploring the domain space in in Causality}

Under the view of causality, exploration be seen as an effort to discover the right causal graph so as to minimize the error in a proposed treatment effect. The most basic algorithm is the IC (Inductive Causation) which to uncovers a causal graph given a fixed distribution $P$, largely by adding single edges to unconnected variables in the causal structure and testing for consistency with observed data. Although IC doesn't include a way to update the experiments producing the distribution $P$, several other algorithms do. Algorithms for discovering causality are broadly based into two categories \citep{mooij2016distinguishing}: 
\begin{itemize}
    \item Conditional independence studies, which are based on the idea that the factorization of $p_{C,E}(c,e)$ for cause $C$ and effect $E$ into $p_{c}(c)p_{E|C}(e|c)$ is lower 
    \item Interventions indicated by the do-calculus (do X = x) described previously
\end{itemize}

 The former often begins with a dump of observational data. While it is difficult to find papers explicitly suggesting how to gather additional observational data after partly constructing a model, we expect, for instance, that in the case of an additive noise model \citep{hoyer2008estimation}, data relating to the relation with the lowest strength would be most helpful.
 
 Within the latter, there are notions of "active" or "adaptive" causal learning, where again the experimenter's next actions in the problem domain are informed by information discovered so far. \citep{hauser2014two} in particular show how to select interventions from optimal coloring of chordal sub-graphs among the variables. The decision of what experiment to perform next is informed by data from previous experiments and constructed model, a mode of operation very similar to the behavior policy in reinforcement learning. 

\subsection{Theoretical limits on the lower bound}
Lastly, we would expect that similar fields solving similar problems would arrive at the same lower bounds of variable error. One way in which that manifests itself is in the Cramer-Rao lower bound. 

The Cramer-Rao lower bound is a bound in the variance of unbiased estimators. The variance of the parameter is going to be as high as its inverse Fisher Information, $var(\theta)\geq\frac{1}{I(\theta)}$, where $I(\theta)$ is a function of the likelyhood function for each of the model parameters. 

In reinforcement learning,  \citep{jiang2015doubly} show that the variance in predictions of doubly robust estimators approach this bound. In causality, backdoor and frontdoor adjustments of treatment effects are methods to use an existing causal model to make predictions. \citep{gupta2020estimating} show that in an over-identified scenario, an optimal estimator approaches the Cramer-Rao lower bound for an information matrix made up of model parameters.

\section{Conclusion}\label{sec:conclusion}
In this paper, we point out that both causal inference and batch RL deal with counterfactual outcomes in a quantitative way, where past experiences are leveraged to make predictions under interventions. With RCM and SCM, causal inference quantitatively measures the effect of a treatment or an intervention, which is of vital importance for planning in unknown domain. As we have discussed in off-policy problems, the difficulty of decision making in unknown environments without online interaction is that poor estimation or limited knowledge of the environment can not be further improved by trial and error. Causality revealed by causal inference shed a new light on this issue, which equips the autonomous agent with better generalization ability, enabling the agent to estimate the counterfactual outcomes and behave accordingly. Furthermore, built on the similarity and close connection between causality and reinforcement learning, the further comparisons between casual approach and RL are made on the subject of data exploration, error estimation and lower-bound limits. 

These investigations in causal inference and batch RL not only are an in-depth study on the interplay between two methodologies for tackling the same problem, but also reveal the advantages of combining these two different approaches for a better and safer RL. A potential future direction we propose would be to incorporate causal inference into a broader range of RL, such as transfer learning and multi-task RL for improving the generalization of RL algorithms, where causality behind data can be transferred among agents in different domains, making reinforcement learning more suitable for real-world applications.

\bibliography{ref}

\end{document}